\documentclass[12pt,fleqn]{article}
\usepackage{xiiiemc}
\usepackage{natbib}
\usepackage{fancyhdr}
\usepackage{color}
\usepackage{wallpaper} 
\usepackage{titlesec}   
\usepackage{float} 	
\usepackage[usenames,dvipsnames]{xcolor}
\usepackage{hyperref} 	
\hypersetup{
	pdfcreator={LaTeX with abnTeX2},
	pdfkeywords={abnt}{latex}{abntex}{USPSC}{trabalho acadêmico}, 
	colorlinks=true,       		
	linkcolor=blue,          	
	citecolor=blue,        		
	filecolor=magenta,      		
	urlcolor=blue,
	allbordercolors=black,
	bookmarksdepth=4
}
\usepackage{tocloft}
\titleformat{\section}
  {\normalfont\bfseries}{\thesection.}{0.5em}{}
 
\renewcommand\thesection{\arabic{section}}

\let\OLDthebibliography\thebibliography
\renewcommand\thebibliography[1]{\OLDthebibliography{#1} \setlength{\parskip}{0pt}\setlength{\itemsep}{0pt plus 0.3ex}}

\title{A SIMPLE TEXT ANALYTICS MODEL TO ASSIST LITERARY CRITICISM: COMPARATIVE APPROACH AND EXAMPLE ON JAMES JOYCE AGAINST SHAKESPEARE AND THE BIBLE} 

\author
    {\rm \begin{tabular}{l} 
    \textbf{Renato Fabbri}$^{1}$ - {\textnormal renato.fabbri@gmail.com}\\%
    \textbf{Luis Henrique Garcia}$^{2}$ - {\textnormal henriquegarcia.pesquisa@gmail.com}\\
    {\fontsize{11}{0}\selectfont $^{1}$University of São Paulo, Institute of Mathematical and Computer Sciences - São Carlos, SP, Brazil}\vspace*{-0.05cm} \\
    {\fontsize{11}{0}\selectfont $^{2}$University of Campinas, Institute of Language Studies - Campinas, SP, Brazil}\vspace*{-0.05cm}\\
  \end{tabular}}

\fancypagestyle{firspagetstyle}
{
	\lhead{}
	\fancyhead[C]{%
		\includegraphics[width=0.9\linewidth]{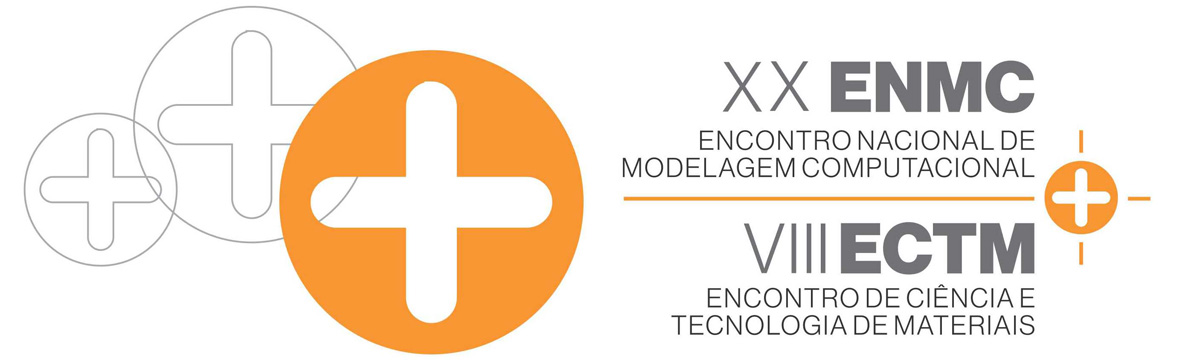}\\%
		{\scriptsize \fontfamily{phv}\fontseries{b}\selectfont \color[rgb]{0.45,0.45,0.45}
		16 a 19 de Outubro de 2017\\
		Instituto Politécnico - Universidade do Estado de Rio de Janeiro\\
		Nova Friburgo - RJ\\
	    }
	}
	\renewcommand{\headrulewidth}{0.0pt}
	\fancyfoot[C]{\footnotesize \parbox{15cm} {\centering  \fontsize{7.5}{0}\selectfont \it Anais do XX ENMC – Encontro Nacional de Modelagem Computacional e VIII ECTM – Encontro de Ciências e Tecnologia de Materiais,  Nova Friburgo, RJ – 16 a 19 Outubro 2017}} 
	\rhead{}
}

\begin{document}
\maketitle

\thispagestyle{firspagetstyle}

\fancyhead[L]{\footnotesize{\fontsize{7.5}{0}\selectfont \it XX ENMC e VIII ECTM\\
	16 a 19 de Outubro de 2017\\
	Instituto Politécnico Universidade do Estado do Rio de Janeiro – Nova Friburgo - RJ\\}}
\renewcommand{\headrulewidth}{0.0pt}
\fancyfoot[C]{\footnotesize \parbox{15cm} {\centering  \fontsize{7.5}{0}\selectfont \it Anais do XX ENMC – Encontro Nacional de Modelagem Computacional e VIII ECTM – Encontro de Ciências e Tecnologia de Materiais,  Nova Friburgo, RJ – 16 a 19 Outubro 2017}} 
\rhead{}

\begin{abstract}
    Literary analysis, criticism or studies is a largely valued field with dedicated journals and researchers
    which remains mostly within the humanities scope.
    Text analytics is the computer-aided process of deriving information from texts.
    In this article we describe a simple and generic model for performing literary analysis
    using text analytics.
    The method relies on statistical measures of: 1) token and sentence sizes and
    2) Wordnet synset features.
    These measures are then used in Principal Component Analysis where the texts to be analyzed
    are observed against Shakespeare and the Bible, regarded as reference literature.
    The model is validated by analyzing selected works from James Joyce (1882-1941),
    one of the most important writers of the 20th century.
    We discuss the consistency of this approach, the reasons why we did not use other
    techniques (e.g. part-of-speech tagging) and the ways by which the analysis model might be adapted and enhanced.
\end{abstract}

\keywords{\em{Text analytics, Literary criticism, Wordnet, Shakespeare, Bible}}

\pagestyle{fancy}

\section{INTRODUCTION}
Literary criticism (also literary criticism or literary studies)
is performed by intellectuals using various techniques,
including intuition and contextualization through erudition~\citep{litCri}.
Text analytics is usually considered a synonym of text mining,
i.e. data mining applied to textual data, the extraction of meaningful information
from texts by means of computer-aided analysis.
A difference can be established nevertheless:
text mining is more associated to earlier applications (e.g. dating to the 1980s)
and to specific tasks, while the term
text analytics is more frequent nowadays and might be related to a
less purposeful processing of textual data.
Accordingly, for example, a word cloud is more easily associated to
text analytics while a search engine is more promptly associated to text mining~\citep{tmWiki}.

In this work we propose a very simple and generic model for literature analysis
by means of statistical measures, Principal Component Analysis (PCA) and comparison against
reference literature.
The uncomplicated methods favor the collaboration between researchers of different backgrounds.
For example: a computer science professional can understand, adapt and expand the techniques
while a literature scholar can deepen the interpretation and assert the relevance of the conclusions.

Section~\ref{sec:matMet} describes the corpus and methods.
Section~\ref{sec:res} is dedicated to the presentation and discussion of results.
Section~\ref{sec:conc} holds conclusions and further work considerations.

\section{MATERIALS AND METHODS}\label{sec:matMet}
\subsection{Corpus}
This work encompasses a comparison of the literature to be analyzed against
reference literature.
What is regarded as reference literature is arbitrary and we chose them,
within this presentation and first formalization, to be
possibly the two greatest references of the English literature~\citep{bib,shake}:

\begin{itemize}
    \item the complete works by William Shakespeare as given by the
        publication in the Gutenberg Project~\citep{shakWhole}:
        36 plays (tragedies, comedies and historical) and poetry (2 batches).
        Shakespeare is often recognized as the greatest writer of the English language
        and is a universal reference of literature.
    \item The 80 books of the King James Bible, including Old Testament (39 books),
        Apocrypha (14 books) and New Testament (27 books).
        This is the most referenced English translation of the Bible.
        These books are also universally accredited for their influence in English literature.
\end{itemize}

We should emphasize that changing this reference literature is very straightforward.
One should only provide the corresponding text files and modify the scripts to read the
intended records.
If the works are well-known, the process should require only a quick search on the web
(e.g. within Gutenberg or Archive.org projects), saving the text locally and then changing
filenames in the scripts.
Some possibilities include: other masters of English literature;
a selection of poets; works from scientific literature of a specific field;
works in other language, such as Machado de Assis and Clarice Lispector
if analyzing works in Brazilian Portuguese.
There is no reason why the corpus should not include data streaming (e.g. from Twitter)
or access to online resources,
such as Wikipedia pages.

To illustrate and validate the method,
we performed and herein report an analysis of a selection of works written by James Joyce:
\begin{itemize}
    \item Stephen Hero: written around 1905 and published posthumously in 1944, an autobiographical novel of which part is lost
        (Joyce threw it on fire after a number of rejections by publishers).
    \item Dubliners: published in 1914, it is a collection of 15 short stories about Dublin's middle class.
    \item A Portrait of the Artist as a Young Man: published in 1916, a condensed and reworked version of Stephen Hero. 
    \item Ulysses: published in 1922, considered one of the most important works of the modernist literature.
    \item Finnegans Wake: published in 1939, often considered one of the most difficult fictional works of the English language,
        the last work written by Joyce.
\end{itemize}

\subsection{Pre-processing}
The reference literature (Shakespeare and Bible books) were cleaned and separated into
individual files.
As both collections do not hold well defined paragraph structures,
these were discarded.
These routines can be inspected through reading the scripts in Table~\ref{tab:files}.

\begin{table}[H] 
	\caption{Files related to the analysis model proposed in this article.
	All files are found in a public git repository~\citep{repo}.}\label{tab:files}
\vspace{12pt}
\centering{}
	\begin{tabular}{  c | p{7cm} }
	\textbf{File}           & \textbf{Description} \\\hline
	\texttt{scripts/analysis.py}  & Python script that makes the initial quantification of the books. \\
	\texttt{scripts/analysis2.py}  & Python script that performs PCA and renders the figures with scatter plots. \\
	\texttt{scripts/BibleSeparation.py}  & Python scripts that separates the King James Bible text into files with individual books. \\
	\texttt{scripts/shakespeareSeparation.py}  & Python scripts that separates the text with the complete works of William Shakespeare into files with individual books. \\
        \texttt{corpus/*}  & Text files corresponding to individual books from Shakespeare, Joyce and the Bible.  \\
	\texttt{latex/*}  & The PDF of this article and the files necessary to render it. It is the main documentation of the proposed analysis model. \\
\end{tabular}
\end{table}

\subsection{Analysis routine}
As modeled until the moment, the analysis is performed by: the achievement of meaningful sets of textual elements,
quantifying their incidences, taking overall measurements of these quantifications in each of the books,  
performing PCA of the books in the measurements space, plotting the books within principal components
and measures of particular interest, interpreting the results.
We should look at each of these phases:
\begin{itemize}
    \item Achievement of meaningful sets of textual elements: the original texts were separated into sets of: sentences, tokens, stopwords, known words (which are not stopwords), punctuations, tokens which are not stopwords or punctuation, Wordnet~\citep{wordnet} synsets of each known word.
    \item Quantization: each of the sets above were quantified by the mean of their sizes in number of characters of each element, or by means of the number of elements they contain, or by means of synset characteristics (only depth was used in the example analysis).
    \item For the PCA, all the books were considered together. The z-score of each dimension (measure type) was performed to avoid meaningless prevalence of some measures over others (z-score of measures $x_i$ is $x'_i = \frac{x_i-\mu(x)}{\sigma(x)}$ where $\mu(x)$ is the mean of all $x_i$ and $\sigma(x)$ is the standard deviation). Then PCA was performed as usual: performing the eigendecomposition of the covariance matrix (where entry $m_{ij}$ if the covariance of measures $i$ with measures $j$) and observing the eigenvectors associated to the greatest eigenvalues.
    \item For visual inspection of the resulting structures, we used scatter plots of principal components and of measures which were relevant to our analysis of James Joyce's works.
    \item For the interpretation of the results, we made discussions about literary criticism and bout the analysis of James Joyce before performing the quantitative analysis described above.
        When the final figures and measures were done, we had another round of considerations about what they revealed.
\end{itemize}

We discarded using other techniques mainly because of three reasons:
1) other methods involve greater complexity and would not favor the communication between interested parties;
2) other methods might not be so easily applied to generic texts,
e.g. part-of-speech tagging relies heavily on the vocabulary and the syntactic structure
which are used with deep innovations by literary authors, especially from the start of last century and thereon;
3) using only the measures mentioned above, we already reached 20 dimensions.
Nevertheless, we encourage adapting the method by inclusion of other measures and of other analysis procedures beyond PCA,
and we will probably do so in further considerations of this endeavor.

\section{RESULTS AND DISCUSSION}\label{sec:res}

\begin{figure}[!htbp] 
\vspace{-2pt}
\begin{center}
\includegraphics[height=6.7cm,width=9cm]{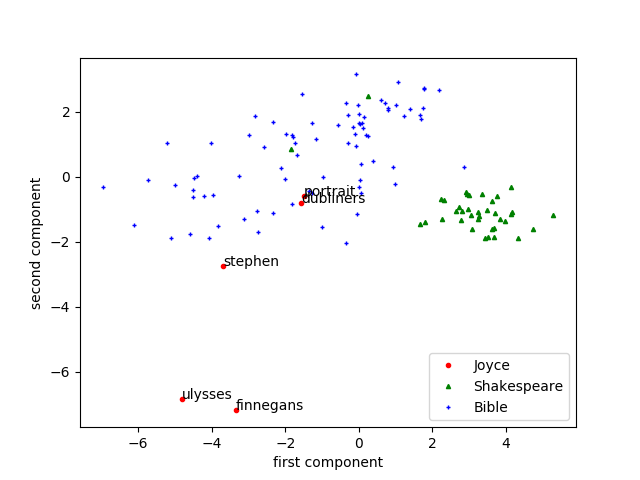}%
    \caption{Reference works (from Shakespeare and the Bible) and a selection of books by James Joyce plotted with respect to the first two principal components.
    As expected, Ulysses and Finnegans Wake are near each other and farther from the other books. Dubliners and Portrait are also near each other but also near other books (in this case, books from the Bible), which is in accordance with their style.
    Stephen Hero is between these two groups, also coherent with expectations.
    There is no prevalence of few measures in these components, reason why we omit this aspect of the analysis.
    The interested reader should access the scripts described in Table~\ref{tab:files} to deepen the analysis exposed here only by way of illustration of the proposed method.}
\label{fig:pca1}%
\end{center}
\end{figure}

\begin{figure}[!htbp] 
\vspace{-2pt}
\begin{center}
\includegraphics[height=6.7cm,width=9cm]{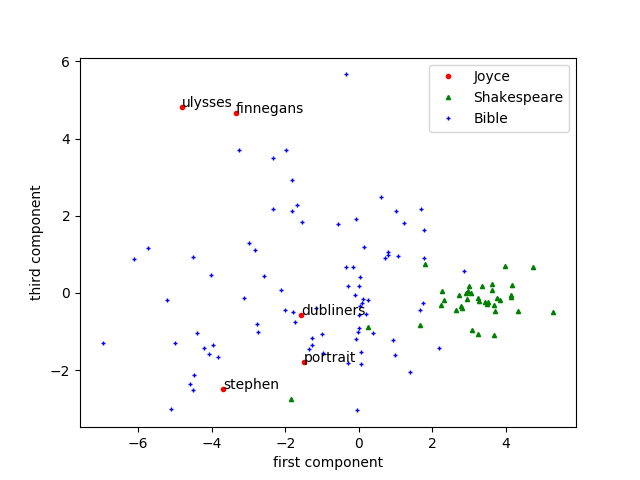}%
\caption{First and third principal components.
    In this case, the separation of the works are less clear.
    This is especially true for Stephen Hero.}
\label{fig:pca2}%
\end{center}
\end{figure}

\begin{figure}[!htbp] 
\vspace{-2pt}
\begin{center}
\includegraphics[height=6.7cm,width=9cm]{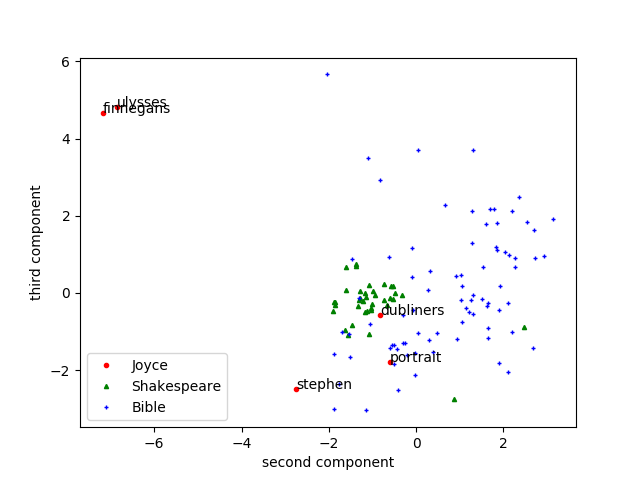}%
\caption{Second and third principal components.
    This is a notable case because it suggests the
    same conclusions as Figure~\ref{fig:pca1} but is even more explicit.
    On one hand, this graph might be regarded as less meaningful than the other because it is related to less relevant components.
    On the other hand, we are analyzing art and more subtle artifices might be the focus of the artist,
    a researcher, or the way the resulting literature is absorbed by the reader.}
\label{fig:pca3}%
\end{center}
\end{figure}

Figures~\ref{fig:pca1},~\ref{fig:pca2} and~\ref{fig:pca3} exposes the works of Joyce and
Shakespeare and the Bible books within the principal components.
As the second and third component held near spreads\footnote{The
amount of dispersion in each component is:
$\approx 40.25,\; 14.32,\; 11.91\%$ and then values bellow $9\%$.} (absolute values of the corresponding eigenvalues),
and the first two components summed only $\approx 50\%$ of all dispersion,
we chose to use the first three components (in contrast to using only the first two components as is usual for PCA).
As can be noticed, Joyce's works are very distinct from Shakespeare,
and some of them are also very distinct from Bible books.
Nevertheless, some of them fall near Bible books.

\begin{figure}[!htbp] 
\vspace{-2pt}
\begin{center}
\includegraphics[height=6.7cm,width=9cm]{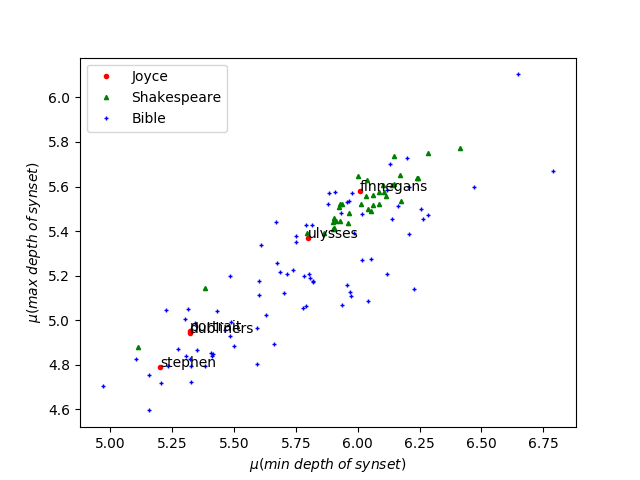}%
\caption{Synset depths. Lower depth is regarded here as evidence of abstraction. 
    In this case, surprisingly, Ulysses and Finnegans Wake are near Shakespeare and have more deeper synsets.
    In other words, it does not reflect the abstraction of the language as we hypothesized before performing the analysis.
    This might mean that we should update our conceptualizations but might also be a byproduct of the fact that these works hold less known words (see Figure~\ref{fig:abst2}), and the ones that are known are used to deploy very definite meaning (i.e. words with deep synsets).}
\label{fig:abst1}%
\end{center}
\end{figure}

\begin{figure}[!htbp] 
\vspace{-2pt}
\begin{center}
\includegraphics[height=6.7cm,width=9cm]{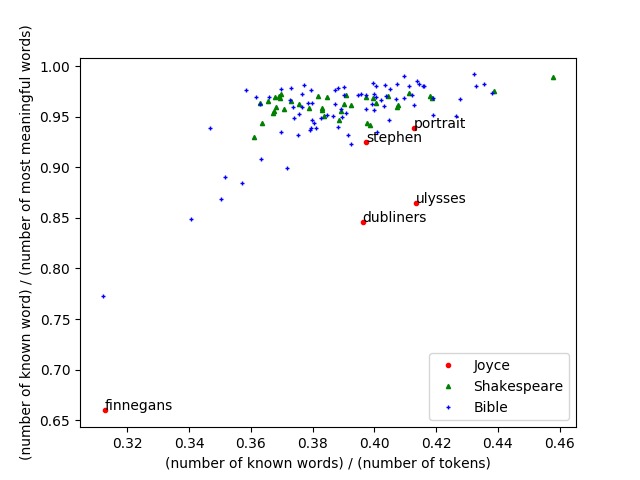}%
    \caption{Fraction of known words among all tokens and among most significant words (words which are not stopwords).
    Lower fraction is regarded here as evidence of abstraction because the reader should infer meaning.
    Finnegans Wake is very distinct from all the books, as expected.
    It is surprising that: 1) Ulysses has a higher rate of known words that Dubliners;
    and 2) that these two measures are the best for a classifier to identify these works by James Joyce, among all the measures used in the figures of this article, including the principal components.}
\label{fig:abst2}%
\end{center}
\end{figure}

Figures~\ref{fig:abst1} and~\ref{fig:abst2} are direct plots of measures
we idealized to probe the extent of the need of interpretation by the reader.
The first plot is dedicated to synset depth\footnote{The
depth of the synset is the number of steps needed to reach the most generic concept~\citep{wordnet}.
For nouns, the most generic concept is "thing".
The max depth is the maximum number of steps while the min depth is the minimum number
of steps.
The tree yielded by the relation of more and less generic concepts (e.g. mammal and horse)
is the "taxonomic tree",
which holds relations of hypernymy/hyponymy or superclass/subclass.}.
The lower the depth, the more abstract the concept is regarded by our analysis.
In this plot, we conclude that three of the works by Joyce lie on
the more abstract margin among the reference works,
but two of them lie within the middle and the more concrete (less abstract)
books.
The second of these plots is dedicated to the amount of unknown words,
and the conclusion is that some of the works have an very distinctive amount
of unknown words, but all of them fall on the greater amount of unknown words
among the most meaningful tokens when the same rate of unknown words among all tokens is considered.

We propose to validate and illustrate the analysis model by considering the
works by Joyce, but, as this is the first work of the kind which analyzes
Shakespeare and the Bible, as far as the authors know,
some considerations about them are also opportune.
First, the works by Shakespeare lie in a notably more restricted domain
when compared against the Bible.
Second, they are perfectly distinguishable with respect to the first two
principal components: a simple Bayesian inference or neural network
should be able to correctly classify a book from one group or the other.
Third, Shakespeare uses a less abstract language at least in the sense captured
by the depth of the synsets.
This diversity is convenient for a reference literature to compare something against.

Finally,
we believe to have reached a good result in terms of the
model proposed for the analysis.
The model is very simple, which favors both elaboration of
variants and the understanding by interested researchers which are potentially from
diverse and multidisciplinary backgrounds.
It is robust, in the sense that it does not rely on canonical vocabulary or syntactic structures.
Furthermore, the method is very fast:
pre-processing and then processing and rendering the figures can all be performed in 
a few minutes.

\section{CONCLUSIONS AND FUTURE WORK}\label{sec:conc}
The analysis model proposed yields interesting results for literary criticism.
It is robust, easily adaptable and fast.
Also, the online availability of the scripts and the reference corpus,
all in public domain, facilitates reuse and the achievement of derivatives.
The example analysis
revealed distinctive traces of the works by James Joyce and can be used to argue quantitatively
in favor of the thesis that the style of Joyce calls the reader to fill the
meaning gaps generated by the abstraction.

In further efforts, we should:
\begin{itemize}
    \item Deepen the analysis of the reference literature (books by Shakespeare and in the Bible)
        to better contextualize any literature we consider against them.
    \item Expand the use of Wordnet to encompass synonymy, antonymy, meronymy, etc.
        Also to consider specific roots of nouns, adjectives, verbs and adverbs.
    \item Report this endeavor to the literary criticism academic community.
        This should be done at least in two ways: by describing the method and its relevance
        within the humanities background;
        and by exposing results from analyzing specific authors, such as Joyce and Ezra Pound.
    \item Consider other measures of abstraction.
	    Should we regard the length of words and sentences as cues of an author's style?
        Should we count the root synsets instead of the depth?
    \item Vary the methods and state reasonable generic bounds e.g. for splitting a work to
        obtain more data points.
    \item Investigate the results exposed in Figure~\ref{fig:abst1} which are not in consonance with what we expected.
    \item Investigate the very unexpected result that Dubliners has more unknown words that Ulysses.
	    This might be an indicative e.g. that in Dubliners the neologisms are more subtle.
		But his will entail an article about text analytics an Joyce, not about an analysis model.
\end{itemize}

\subsection*{\textit{Acknowledgements}}
The authors thank the open source software developers,
especially those who enabled this work by developing
the Python language, Numpy, Matplotlib and the NLTK;
the open culture movement, especially the collaborators
of the Gutenberg and the Archive.org projects, which
enabled this work by making the literature available;
the IFSC/USP, ICMC/USP and IEL/UNICAMP researchers for their attentive
collaboration whenever we required opinions and directions for learning and researching.








\end{document}